\documentclass[acmtog]{acmart}
\acmSubmissionID{}

\usepackage{graphicx}
\usepackage{amsmath}
\usepackage{booktabs}
\usepackage{balance}
\usepackage{acronym} 
\usepackage{comment}
\usepackage{tabularx}
\usepackage{cleveref}

\newcommand{\modelname}{SHELLS\xspace}
\newcommand{\modelnamelong}{\modelname~(Semantic Head Estimation via Layered Local Sampling)\xspace}

\usepackage{booktabs} 

\usepackage[ruled]{algorithm2e} 
\usepackage{booktabs} 
\usepackage{multirow} 
\usepackage{colortbl}
\usepackage{balance}

\SetAlFnt{\small}
\SetAlCapFnt{\small}
\SetAlCapNameFnt{\small}
\SetAlCapHSkip{0pt}

\acmJournal{TOG}

\copyrightyear{2026}
\acmYear{2026}
\setcopyright{cc}
\setcctype{by}
\acmConference[SIGGRAPH Conference Papers '26]{Special Interest Group on Computer Graphics and Interactive Techniques Conference Conference Papers}{July 19--23, 2026}{Los Angeles, CA, USA}
\acmBooktitle{Special Interest Group on Computer Graphics and Interactive Techniques Conference Conference Papers (SIGGRAPH Conference Papers '26), July 19--23, 2026, Los Angeles, CA, USA}
\acmDOI{10.1145/3799902.3811201}
\acmISBN{979-8-4007-2554-8/2026/07}

\begin{document}

\title{Topologically Consistent Multi-view 3D Head Reconstruction via Coarse-Guided Layered Surface Sampling}

\author{Timo Bolkart}
\orcid{0000-0002-3829-3924}
\affiliation{%
 \institution{Google}
 \country{Switzerland}
 }
\email{tbolkart@google.com}

\author{Daoye Wang}
\orcid{0000-0002-2879-6114}
\affiliation{%
 \institution{Google}
 \country{Switzerland}
 }
\email{daoye@google.com}

\author{Prashanth Chandran}
\orcid{0000-0001-6821-5815}
\affiliation{%
 \institution{Google}
 \country{Switzerland}
 }
\email{prchandran@google.com}

\renewcommand\shortauthors{Bolkart et al.}
\newcommand{\R}[1]{\mathbb{R}^{#1}}
\newcommand{\inR}[1]{\in \R{#1}}
\newcommand{\norm}[1]{\left\|#1\right\|}

\newcommand{\scalar}[1]{\lowercase{#1}}
\renewcommand{\vector}[1]{\textbf{\lowercase{#1}}}
\renewcommand{\matrix}[1]{\textbf{\uppercase{#1}}}
\newcommand{\tensor}[1]{\mathcal{\uppercase{#1}}}
\newcommand{\mappingfunction}[1]{\uppercase{#1}}
\newcommand{\misc}[1]{\uppercase{#1}}

\newcommand{\image}{\tensor{I}}
\newcommand{\imagewidth}{\scalar{w}_i}
\newcommand{\imageheight}{\scalar{h}_i}
\newcommand{\numviews}{\scalar{n}_i}

\newcommand{\cameracenter}{\vector{c}}
\newcommand{\cameraparameters}{\misc{C}}
\newcommand{\cameraprojection}{\Pi}

\newcommand{\lorarank}{\scalar{r}}
\newcommand{\featuredim}{\scalar{d}_f}
\newcommand{\fusedfeaturedim}{2\featuredim}
\newcommand{\featureimage}{\tensor{F}}
\newcommand{\featurevector}{\vector{f}}
\newcommand{\featureimagewidth}{\scalar{w}_f}
\newcommand{\featureimageheight}{\scalar{h}_f}

\newcommand{\featuremean}{\boldsymbol{\mu}}
\newcommand{\featurevariance}{\boldsymbol{\sigma}^2}

\newcommand{\numgraphpoints}{\scalar{n}_g}
\newcommand{\numshellpoints}{3\numverticescoarse}
\newcommand{\samplingshelldistance}{\scalar{d}_l}
\newcommand{\samplinggraph}{\matrix{S}_g}
\newcommand{\samplingshells}{\matrix{S}_l}
\newcommand{\samplingshellstemplate}{\matrix{S}_t}
\newcommand{\samplingpoint}{\vector{s}}

\newcommand{\graphfeaturepointcloud}{\matrix{F}_g}
\newcommand{\shellsfeaturepointcloud}{\matrix{F}_l}

\newcommand{\mesh}{\misc{M}}
\newcommand{\meshfinal}{\misc{M}_f}
\newcommand{\meshcoarse}{\hat{\misc{M}}_c}
\newcommand{\meshtemplate}{\misc{M}_t}
\newcommand{\meshtemplatedown}{\hat{\misc{M}}_t}

\newcommand{\meshcoarsenormals}{\hat{\matrix{N}}_c}
\newcommand{\meshgtnormals}{\matrix{N}_\text{gt}}

\newcommand{\numvertices}{\scalar{n}_v}
\newcommand{\numverticescoarse}{\scalar{n}_c}

\newcommand{\viewdir}{\vector{d}}
\newcommand{\vertex}{\vector{v}}
\newcommand{\vertexnormal}{\vector{n}}
\newcommand{\vertices}{\matrix{V}}
\newcommand{\verticesgt}{\matrix{V}_\text{gt}}
\newcommand{\verticescoarse}{\hat{\matrix{V}}_c}
\newcommand{\verticescoarseup}{\matrix{V}_c}
\newcommand{\verticesfinal}{\matrix{V}_f}

\newcommand{\verticescoarsedistance}{\boldsymbol{\Delta}\matrix{V}_c}
\newcommand{\verticesfinaldistance}{\boldsymbol{\Delta}\matrix{V}_f}

\newcommand{\verticestemplate}{\matrix{V}_t}
\newcommand{\verticestemplatedown}{\hat{\matrix{V}}_t}

\newcommand{\faces}{\matrix{T}}
\newcommand{\facesdown}{\hat{\matrix{T}}}

\newcommand{\templatetokens}{\matrix{Z}_t}
\newcommand{\templatetokensdown}{\hat{\matrix{Z}}_t}

\newcommand{\featuretokensgraph}{\matrix{Z}_g}
\newcommand{\featuretokensshells}{\matrix{Z}_l}

\newcommand{\latenttokenscoarse}{\matrix{Z}_c}
\newcommand{\latenttokensfinal}{\matrix{Z}_f}

\newcommand{\querytokenscoarse}{\matrix{Q}_c}
\newcommand{\keytokenscoarse}{\matrix{K}_c}
\newcommand{\querytokensfinal}{\matrix{Q}_f}
\newcommand{\keytokensfinal}{\matrix{K}_f}

\newcommand{\downsamplingmatrix}{\matrix{D}}
\newcommand{\upsamplingmatrix}{\matrix{U}}

\newcommand{\modeldim}{\scalar{d}_m}
\newcommand{\featurenet}{\mappingfunction{F}_{\text{img}}}
\newcommand{\transformer}{\mappingfunction{F}_{\text{pred}}}

\newcommand{\featureweights}{\phi}
\newcommand{\vertexvisibility}{\delta}

\newcommand{\loss}{\mathcal{L}}
\newcommand{\lossvertextovertex}{\mathcal{L}_\text{v2v}}
\newcommand{\lossvertextoplane}{\mathcal{L}_\text{v2p}}
\newcommand{\losstotal}{\mathcal{L}_\text{total}}

\newcommand{\vertexweights}{\boldsymbol{\Omega}}
\newcommand{\vertexweight}{\omega}

\newcommand{\lossweightvertextovertex}{\lambda_{\text{v2v}}}
\newcommand{\lossweightvertextoplane}{\lambda_{\text{v2p}}}

\newcommand{\lossweightcoarse}{\lambda_{c}}
\newcommand{\lossweightfinal}{\lambda_{f}}

\newcommand{\realdata}{$D_{\text{real}}$}
\newcommand{\syntheticdata}{$D_{\text{syn}}$}

\begin{abstract}

We present \modelnamelong, an efficient feed-forward framework for 3D head reconstruction in dense semantic correspondence from multi-view images.
Existing methods typically refine vertices independently via localized feature volumes.
This approach couples memory-intensive feature sampling to mesh resolution, which limits scalability for dense topologies ($\geq 10k$ vertices) and introduces surface noise. 
In contrast, \modelname decouples feature extraction from mesh resolution via a hierarchical sampling strategy.
We extract multi-view features using a DINOv2 backbone with LoRA adaptation, projectively sample a sparse global feature cloud, and predict an intermediate coarse mesh.
This coarse prior guides the construction of layered, surface-aware sampling shells that serve as a discrete search space for the final reconstruction.
\modelname maintains surface consistency while using $88\%$ less inference GPU memory ($\sim2.4$GB vs. $\sim20$GB) than volumetric baselines.
It reduces median registration error by $21\%$ to $29\%$ with a $3.5\times$ inference speedup ($0.08$s vs. $0.29$s) for 18k-vertex meshes.
Notably, our model is trained exclusively on synthetic data yet generalizes effectively to real-world captures, eliminating the need for the costly, pre-registered multi-view datasets common in prior work.

\end{abstract}

%
%

%
%


\newcommand{\mainwidth}{0.22222222\textwidth}
\newcommand{\smallwidth}{0.1111111\textwidth}
\newcommand{\tinywidth}{0.05555555\textwidth}

\begin{teaserfigure}
    \setlength{\tabcolsep}{0pt}
    \renewcommand{\arraystretch}{0} 
    \begin{tabular}{@{}c@{}c@{}c@{}c@{}c@{}}
        \begin{tabular}{c}
            \includegraphics[width=\mainwidth]{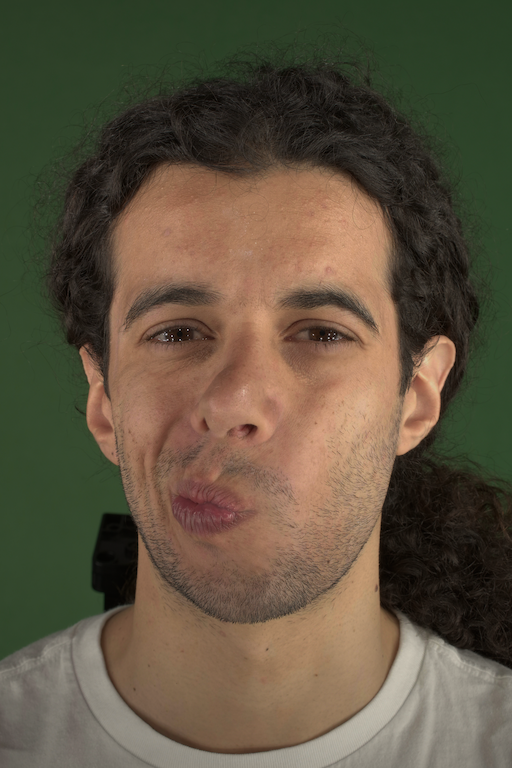}
        \end{tabular} &
        \begin{tabular}{c}
            \includegraphics[width=\tinywidth]{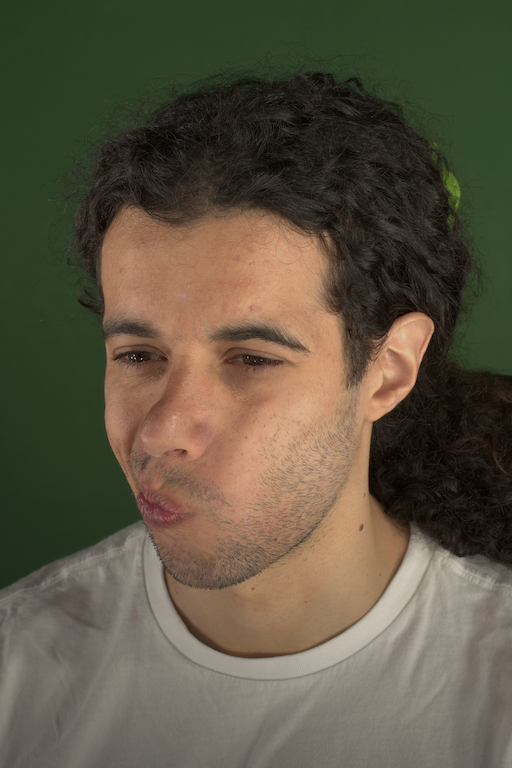} \\
            \includegraphics[width=\tinywidth]{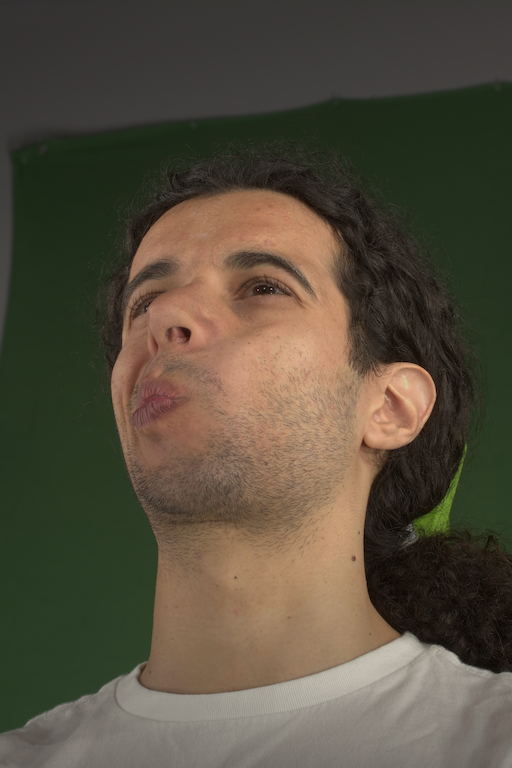} \\
            \includegraphics[width=\tinywidth]{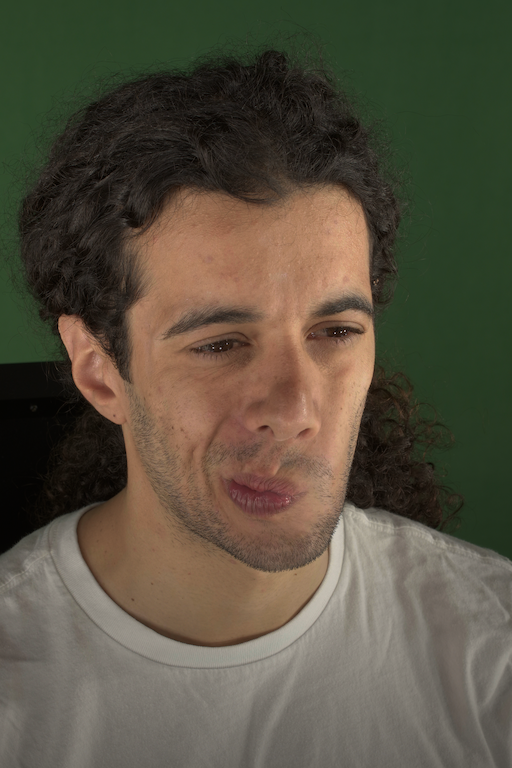} \\
            \includegraphics[width=\tinywidth]{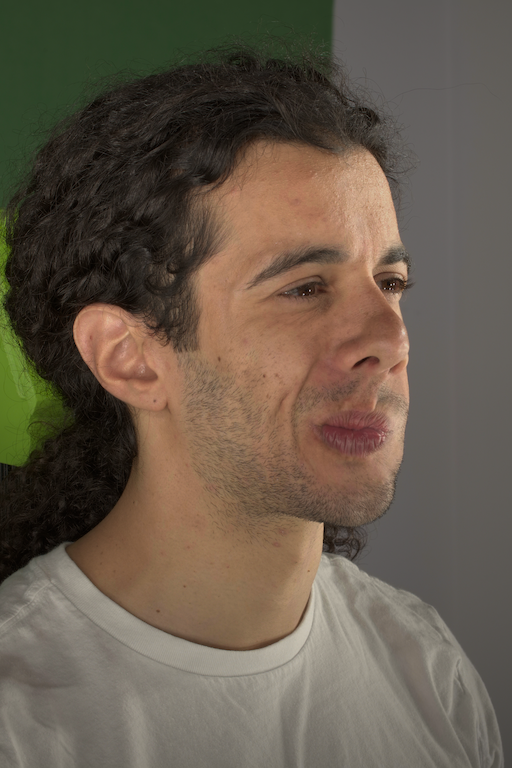}
        \end{tabular} &
        \begin{tabular}{c}
            \includegraphics[width=\mainwidth]{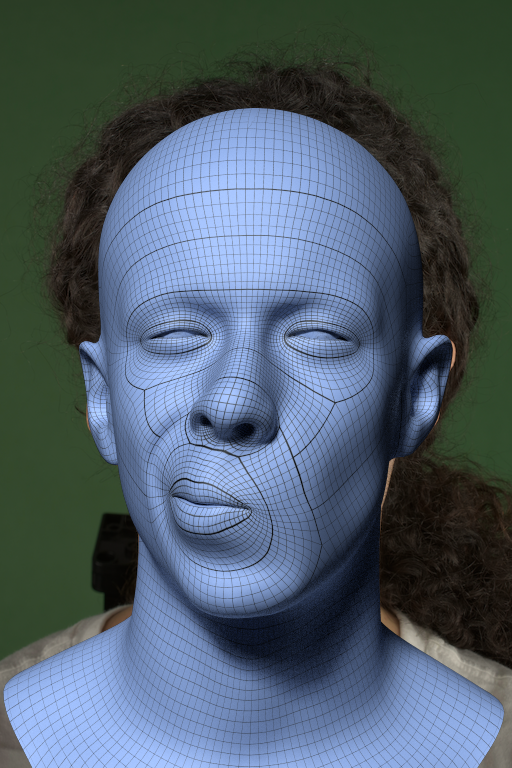}
        \end{tabular} &
        \begin{tabular}{c}
            \includegraphics[width=\tinywidth]{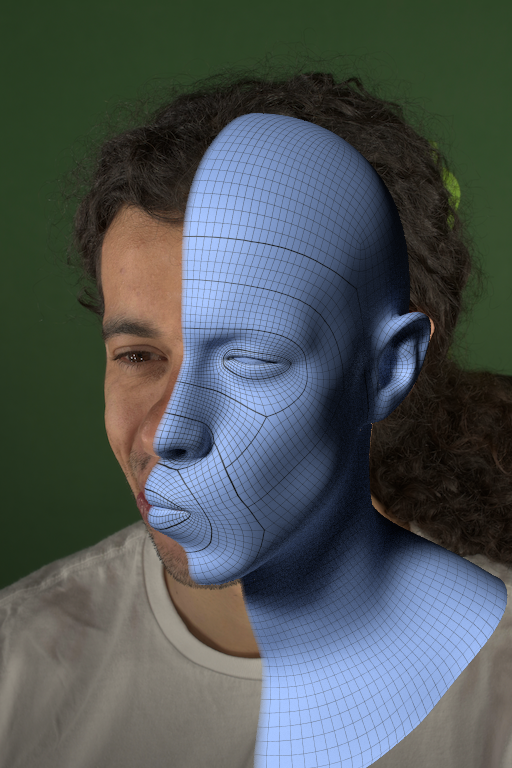} \\
            \includegraphics[width=\tinywidth]{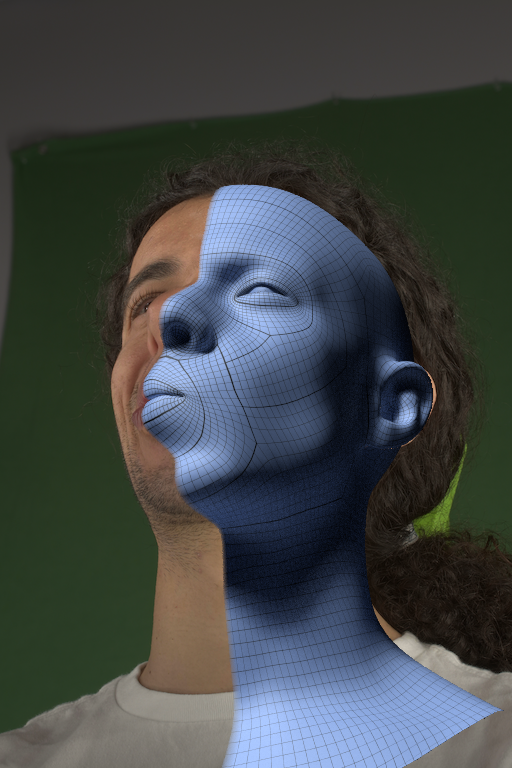} \\
            \includegraphics[width=\tinywidth]{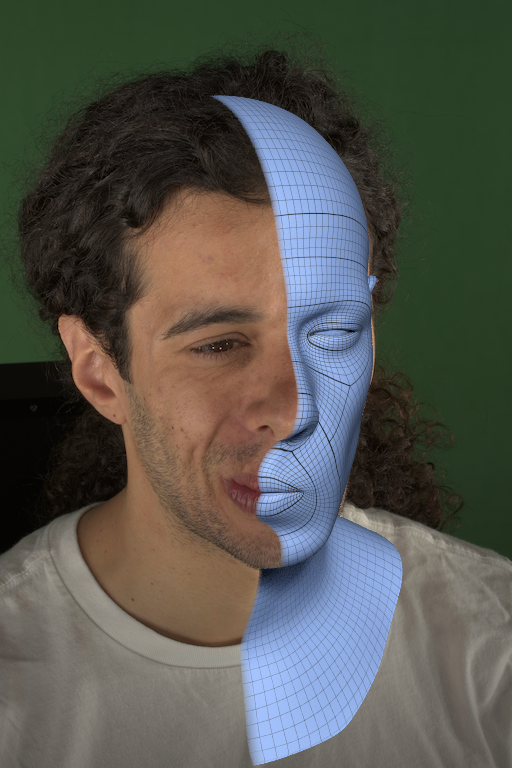} \\
            \includegraphics[width=\tinywidth]{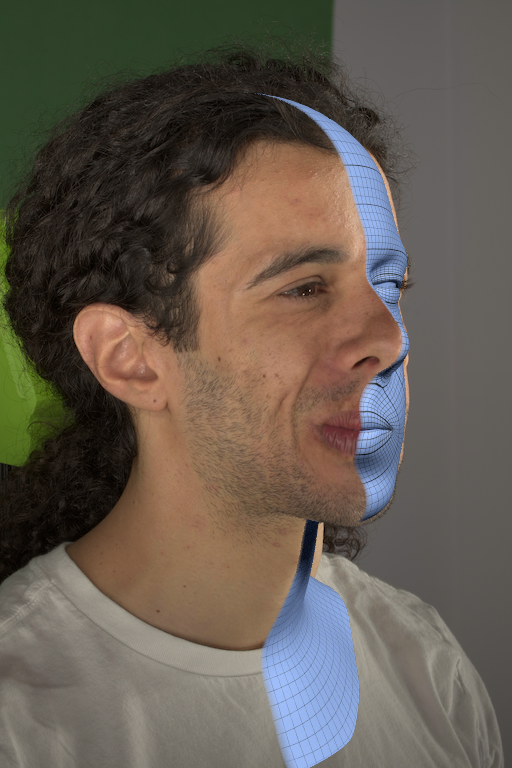}
        \end{tabular} &
        \begin{tabular}{cccc}
            \includegraphics[width=\smallwidth]{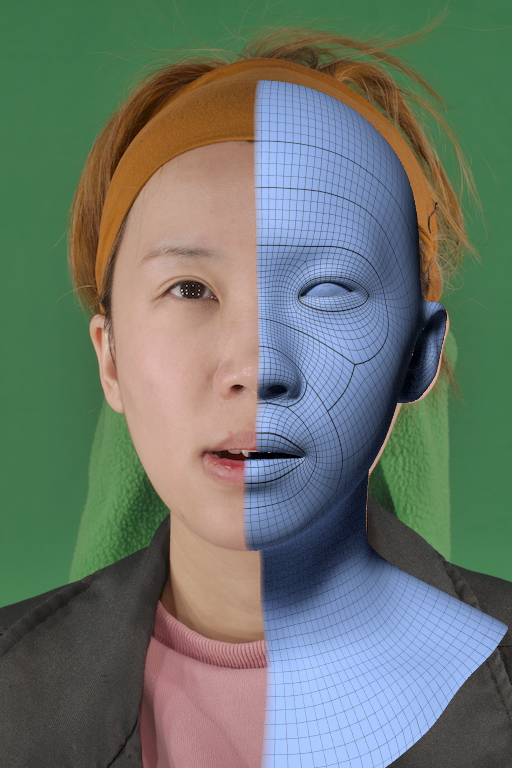} &
            \includegraphics[width=\smallwidth]{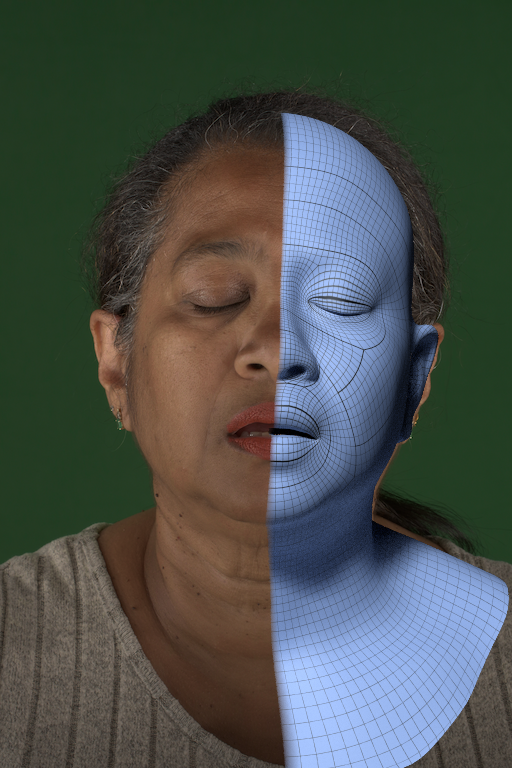} &
            \includegraphics[width=\smallwidth]{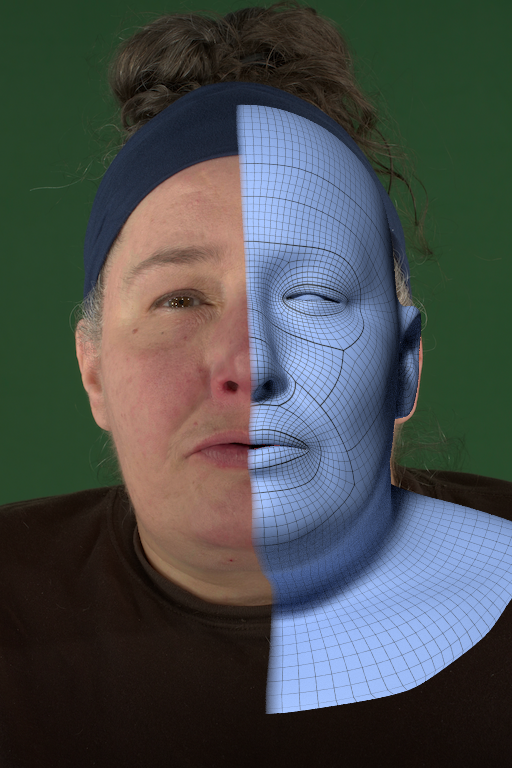} &
            \includegraphics[width=\smallwidth]{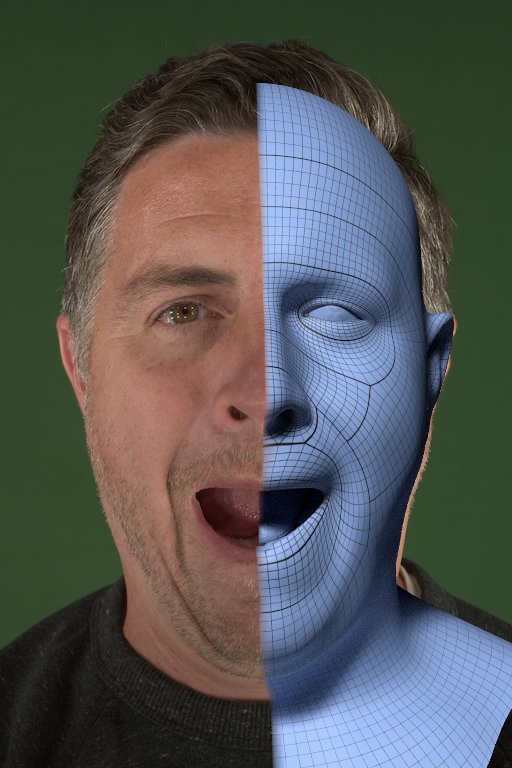} \\
            \includegraphics[width=\smallwidth]{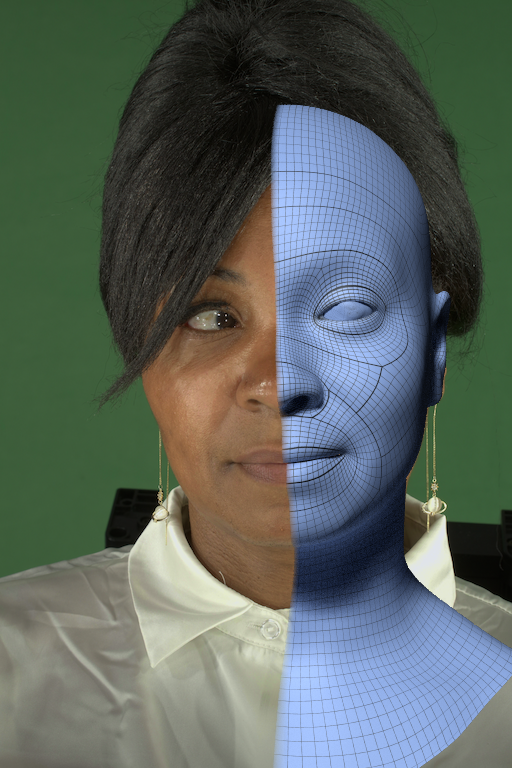} &
            \includegraphics[width=\smallwidth]{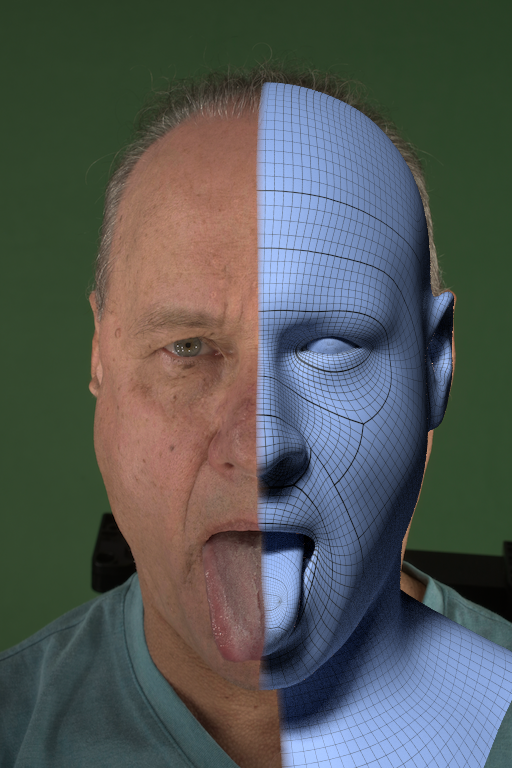} &
            \includegraphics[width=\smallwidth]{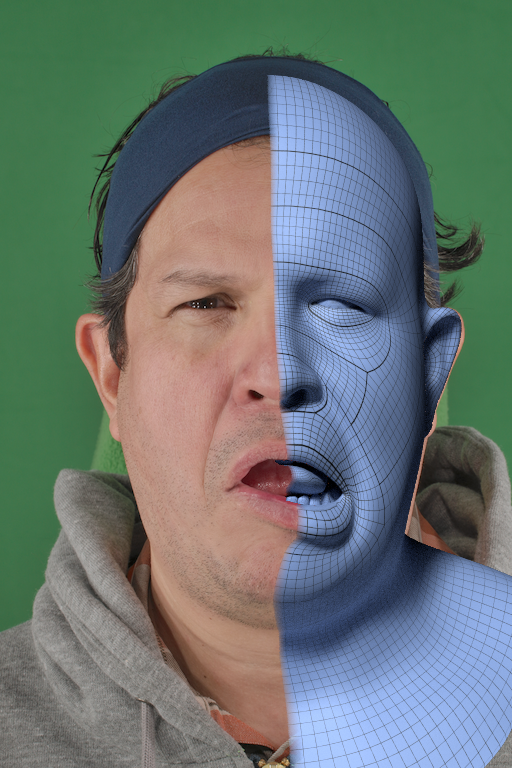} &
            \includegraphics[width=\smallwidth]{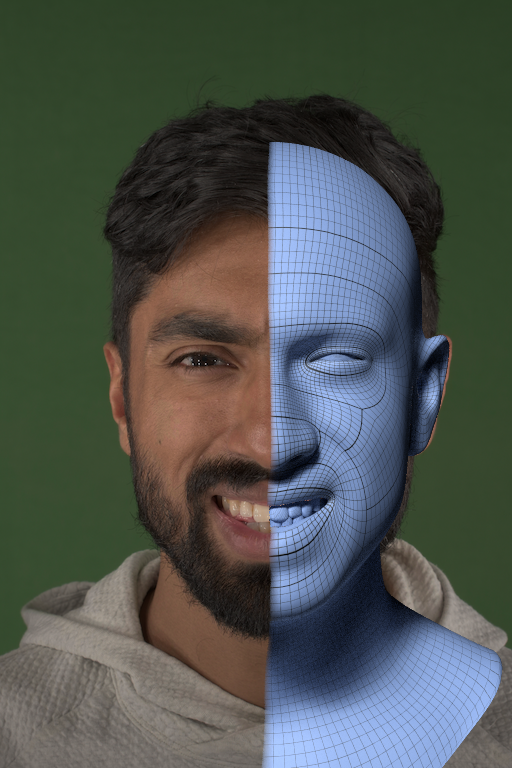}
        \end{tabular}
    \end{tabular}
    \hfill
    \caption{
        \textbf{Feed-forward registration.}
        Given calibrated multi-view images (left; 5 of 13 views shown), \modelname reconstructs 3D meshes in dense semantic correspondence in $0.08$ seconds.
        Overlaid reconstructions demonstrate precise geometric alignment across diverse subjects and expressions (middle \& right).
        \modelname generalizes from synthetic training to real multi-view captures, enabling efficient, high-quality registration of large-scale datasets.
    }
    \label{fig:teaser}
\end{teaserfigure}

\maketitle

\section{Introduction}
\label{sec:introduction}

High-fidelity 3D head reconstruction in dense semantic correspondence is a fundamental requirement for building realistic digital humans \cite{egger2020survey,zielonka2026star}. 
Traditional pipelines register unstructured multi-view stereo (MVS) scans to a unified topology \cite{beeler2011highquality,egger2020survey}.
However, MVS often produces noise and holes in specular regions, necessitating labor-intensive manual cleanup.
It also tends to over-smooth concavities like the ears or nostrils and produces unrealistic geometry in hair regions. 
The subsequent registration step is computationally exhaustive, often requiring several minutes to hours of optimization per frame, and labor-intense as it requires manual clean up and hyperparameter tuning to balance surface fidelity against robustness to scan artifacts \cite{alexander2009emily,seymour2017meetmike}.

To bypass these bottlenecks, recent learning-based frameworks \cite{li2021tofu, bolkart2023tempeh, li2024grape, liu2022refa, filntisis2026mochi} move toward direct surface regressions from calibrated multi-view images.
By eliminating the MVS and registration steps, these methods achieve near-interactive reconstruction speeds, demonstrating great potential for the fully automated processing of large-scale datasets. 
However, while these methods offer superior scalability, they still lack the geometric details of classical registration and face significant architectural bottlenecks. 
State-of-the-art methods \cite{li2021tofu,bolkart2023tempeh,li2024grape} rely on memory-intense global and per-point feature volumes, which constrain the output resolution to low vertex counts ($\sim 3k$ to $\sim 5k$).

To overcome these challenges, we present \modelname, a transformer-based framework that predicts higher-resolution 3D heads in dense semantic correspondence from calibrated multi-view images.
Our approach builds upon ToFu's \cite{li2021tofu} projective multi-view feature sampling strategy, which naturally integrates known camera parameters. 
However, we replace their memory-intense dense feature volumes with a coarse-guided hierarchical feature sampling strategy. 
Specifically, \modelname first employs a sparse global sampling graph to estimate an intermediate low-resolution mesh, which then guides the placement of layered sampling shells displaced along the surface normals.
This surface-aware strategy ensures that feature sampling is restricted to the proximity of the target geometry, reducing the sampling of irrelevant features, and effectively decoupling memory consumption from the final mesh resolution.

Furthermore, existing methods~\cite{li2021tofu,bolkart2023tempeh,li2024grape} lack a global geometric understanding because they refine vertex positions independently, leading to mesh artifacts in regions occluded by hair or clothing.
Our transformer-based prediction model instead processes the sampled features holistically, which improves robustness in the presence of severe occlusions. 
Finally, \modelname eliminates the need for costly real-world data processing.
While prior work requires paired capture data with registration meshes per frame \cite{li2021tofu, liu2022refa} or raw scans for joint optimization \cite{bolkart2023tempeh, li2024grape}, we demonstrate that training \modelname exclusively on synthetic multi-view data is sufficient to generalize to real-world captures (see Fig.~\ref{fig:teaser}).

In summary, we introduce a hierarchical shell-based sampling strategy that reduces GPU memory requirements by $70\%$ for training ($\sim20$GB vs. $\sim65$GB) and $88\%$ for inference ($\sim2.4$GB vs. $\sim20$GB) compared to volumetric baselines.
Our transformer-based architecture holistically predicts 3D faces in dense semantic correspondence, with a $21\%-29\%$ lower median registration error on real capture test data and synthetic test data.
We demonstrate that training on synthetic data is sufficient to generalize to real multi-view captures. 

\section{Related work}
\label{sec:related_work}

\paragraph{Optimization-based registration.}
Traditional face registration often deforms a template mesh non-rigidly to multi-view stereo (MVS) scans \cite{egger2020survey}.
These techniques have matured from neutral expressions \cite{blanz19993dmm} to high-quality performance sequences \cite{beeler2011highquality} and the automatic processing of thousands of identities \cite{booth201610000faces} and sequences \cite{li2017flame}. 
However, MVS reliance introduces noise and holes that necessitate computationally expensive, manually tuned regularization.
Direct methods maximize consistency via differentiable rasterization  \cite{qian2024vhap, qian2024gaussianavatars}, optical flow \cite{Fyffe2017}, or they integrate learnable priors \cite{bai2020dfnrmvs}, neural volume rendering \cite{wang2026reconstructing}, or surface-aligned Gaussians \cite{li2024topo4d}.
While these improve fidelity, their iterative nature remains a bottleneck, requiring minutes to hours per frame.
In contrast, \modelname provides dense correspondence in an efficient, feed-forward manner.

\paragraph{Feed-forward mesh prediction.}
Feed-forward registration accelerates inference through direct mesh prediction.
ToFu \cite{li2021tofu} pioneered volumetric feature sampling for sub-second, consistent multi-view face reconstruction.
Building on ToFu, TEMPEH \cite{bolkart2023tempeh} adds head localization and direct scan supervision, GRAPE \cite{li2024grape} incorporates visual hull initialization, and MOCHI \cite{filntisis2026mochi} eliminates registration supervision.
However, these volumetric models rely on memory-heavy sampling grids that scale poorly to dense topologies.
\modelname instead employs sparse hierarchical sampling and holistic reconstruction to improve memory efficiency and surface coherence.

Similar to \modelname, POEM \cite{yang2023poem,yang2025multiviewhand}  utilizes sparse sampling and a transformer but iteratively refines hand meshes.
In contrast, \modelname is non-iterative and uses joint attention to predict the output in a single pass via an attention-weighted sum of sampling coordinates. 
While POEM targets low-resolution MANO \cite{romero2017mano} hands (778 vertices), \modelname reconstructs significantly denser head topologies (18k vertices). 

Finally, unlike unconstrained methods \cite{vhavleJ025camera3dmm,giebenhain2025pixel3dmm}, \modelname leverages calibrated cameras for metrical accuracy and bypasses the expressiveness limits of linear 3DMMs.

\paragraph{Unstructured points prediction.}
Learning-based multi-view reconstruction significantly improves 3D fidelity \cite{gu2020cascadedvolume,im2019dspnet,kar2017learning,sitzmann2019deepvoxels,yao2018mvsnet,qiu2024chosen}.
Transformer models like DUSt3R \cite{wang2024dust3r}, MASt3R \cite{leroy2024mast3r}, and VGGT \cite{wang2025vggt} predict point maps without explicit calibration but produce unstructured outputs lacking semantic labels or consistent topology. 
While VGGT and St4RTrack \cite{feng2025st4track} incorporate tracking, correspondence remains restricted to the sequence level. 
Consequently, performing reconstruction across different subjects fails to provide inter-subject correspondence. 
In contrast, \modelname regresses meshes in a fixed mesh topology to ensure dense semantic correspondence across both time and identities.

\paragraph{Synthetic data training.}
Synthetic head datasets support diverse applications, from 2D landmark prediction \cite{wood2022denselandmarks} and face parsing \cite{wood2021fakeit} to scan segmentation \cite{chen2025pixel2points} and neural avatar construction \cite{saunders2025gasp, zielonka2025synshot}.
This versatility motivates our use of synthetic data for the multi-view prediction task.

\section{Method}
\label{sec:method}

\begin{figure*}[ht]
    \centering
    \includegraphics[width=1.0\textwidth]{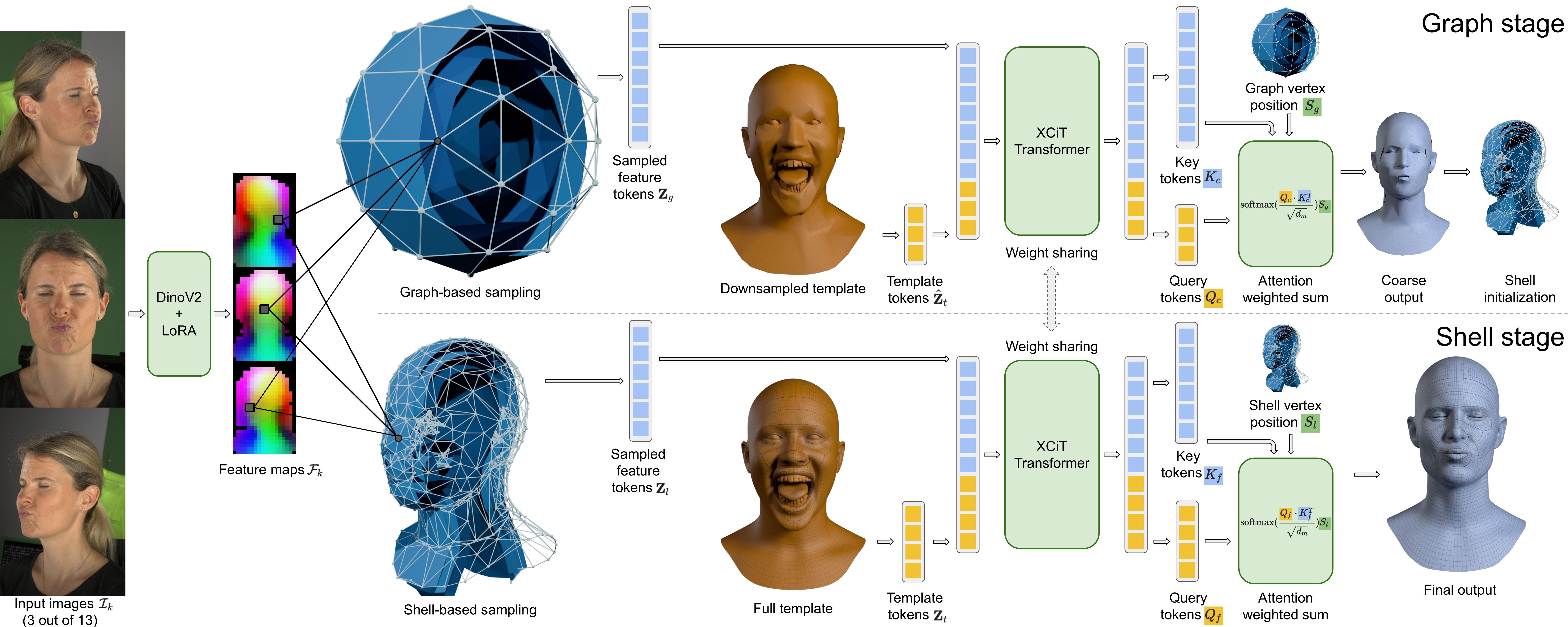}  
    \caption{
        \textbf{Overview of \modelname}. 
        A shared DINOv2 backbone with LoRA adaptation extracts per-view feature maps from the input images (left).
        The graph stage (top) projectively samples features for a sparse graph and processes them alongside a downsampled tokenized template using an XCiT-based transformer.
        From the transformer output, a coarse mesh is regressed as an attention-weighted sum over the sampling graph coordinates.
        This coarse prediction is displaced along its normals to construct sampling shells for surface-aware feature sampling.
        Finally, the shared transformer aggregates these shell-based features with a full-resolution tokenized template to predict the high-fidelity mesh as an attention-weighted sum of dynamic shell coordinates (bottom).
    }
	\label{fig:overview}
\end{figure*}

Given $\numviews$ time-synchronized input images $\{\image_k \inR{\imageheight \times \imagewidth \times 3}\}_{k=1}^{\numviews}$, each of height $\imageheight$ and width $\imagewidth$ and their corresponding camera parameters $\{\cameraparameters_k\}_{k=1}^{\numviews}$ (extrinsics, intrinsics, and lens distortions), \modelname infers a 3D head mesh $\meshfinal :=(\verticesfinal, \faces)$ with vertices $\verticesfinal \inR{\numvertices \times 3}$ in a fixed mesh topology $\faces$.
The fixed mesh topology with a constant number of vertices $\numvertices$ ensures that all reconstructions are in dense semantic correspondence.
As shown in Fig.~\ref{fig:overview}, \modelname consists of two stages, trained end-to-end.
The first stage predicts an intermediate coarse mesh $\meshcoarse := (\verticescoarse, \facesdown )$ with $\numverticescoarse$ vertices $\verticescoarse \inR{\numverticescoarse \times 3}$, which guides feature sampling for the subsequent prediction stage.
The second stage then builds layers of sampling shells around $\meshcoarse$ and predicts the final mesh $\meshfinal$ from the sampled multi-view features.

\paragraph{Feature extraction}
Each image $\image_k$ is processed by a shared feature extraction network $\featurenet(.)$ using a frozen DINOv2~\cite{oquab2023dinov2} backbone to extract 2D feature maps $\featurenet(\image_k) \rightarrow \featureimage_k \inR{\featureimageheight \times \featureimagewidth \times \featuredim}$.
We incorporate trainable LoRA~\cite{hu2022lora} layers with rank $\lorarank$ as residuals in each linear DINOv2 layer to adapt the backbone to the reconstruction task. 
The spatial dimensions are downsampled by a factor of $14$, such that $\featureimageheight=\imageheight/14$ and $\featureimagewidth=\imagewidth/14$.

\subsection{Graph-based coarse prediction}

\paragraph{Feature sampling}

To localize the face within the capture volume when its 3D position is unknown, we employ a sparse sampling strategy.
Instead of utilizing a dense 3D grid~\cite{li2021tofu, bolkart2023tempeh}, we define a sparse point cloud $\samplinggraph \inR{\numgraphpoints \times 3}$ comprising vertices of equidistant concentric spheres, each approximated by a twice-subdivided icosahedron.
For each input image $\{\image_k\}_{k=1}^{\numviews}$, every sampling point $\samplingpoint \inR{3}$ in $\samplinggraph$ is perspectively projected into the image plane using the camera projection $\cameraprojection_k: \mathbb{R}^{3} \rightarrow \mathbb{R}^{2}$ with the given camera parameters $\cameraparameters_k$.
Per-view feature vectors $\featurevector_k \inR{\featuredim}$ are extracted from the feature maps $\featureimage_k$ at the projected 2D locations via bilinear sampling and subsequently fused across all views.

\paragraph{Feature fusion}
Following ToFu~\cite{li2021tofu}, we fuse these per-view vectors by computing the element-wise mean $\featuremean \inR{\featuredim}$ and variance $\featurevariance \inR{\featuredim}$.
The resulting fused feature vector $\featurevector = [\featuremean; \featurevariance] \inR{\fusedfeaturedim}$ is the concatenation of $\featuremean$ and $\featurevariance$.
Performing the feature sampling for all sampling points of $\samplinggraph$ yields the global feature point cloud $\graphfeaturepointcloud \inR{\numgraphpoints \times \fusedfeaturedim}$.

\paragraph{Transformer-based mesh prediction}
We utilize a transformer-based architecture to predict $\verticescoarse$ from the sparse feature point cloud.
We define a fixed template mesh $\meshtemplate := (\verticestemplate, \faces)$, with vertices $\verticestemplate \inR{\numvertices \times 3}$, which establishes the mesh topology of the output prediction.
For computational efficiency, the template mesh is downsampled to a lower resolution $\meshtemplatedown = (\verticestemplatedown, \facesdown)$ with $\numverticescoarse$ vertices using iterative surface simplification~\cite{garland1997simplification}.
Similar to ToFu's~\cite{li2021tofu} coarse stage, this makes the intermediate prediction independent of the final mesh resolution. 

Following \citet{ranjan2018coma}, we derive a downsampling matrix $\downsamplingmatrix \in \{0,1\}^{\numvertices \times \numverticescoarse}$ and an upsampling matrix $\upsamplingmatrix \inR{\numverticescoarse \times \numvertices}$ based on barycentric interpolation.
The coarse template vertices $\verticestemplatedown := \downsamplingmatrix\verticestemplate \inR{\numverticescoarse \times 3}$ are processed by an MLP to generate tokens $\templatetokensdown \inR{\numverticescoarse \times \modeldim}$.

Simultaneously, the coordinates of the fixed sampling graph $\samplinggraph$ are processed through a separate MLP to form a geometric positional embedding. 
This embedding is element-wise added to the multi-view features $\graphfeaturepointcloud$ and projected to $\modeldim$ via a linear layer to produce feature tokens $\featuretokensgraph \inR{\numgraphpoints \times \modeldim}$.

The joint set of tokens $\latenttokenscoarse = [\templatetokensdown; \featuretokensgraph] \inR{(\numverticescoarse+\numgraphpoints) \times \modeldim}$ is processed by a sequence of transformer layers $\transformer$.
To circumvent the quadratic memory complexity of standard self-attention, each layer adopts the Cross-Covariance Image Transformer (XCiT) architecture~\cite{ali2021xcit}, following its successful application to 3D face modeling~\cite{chandran2022shape}.
Utilizing a parallel block design for enhanced efficiency, layer-normalized input tokens are processed concurrently by a cross-covariance attention (XCA) layer and a feed-forward network (FFN). 
By computing attention across the feature dimension rather than the token count, the XCA layer significantly reduces computational complexity when processing large point sets. 
The outputs of these XCA and FFN branches are summed and integrated with the original input via a residual connection.

The transformer output $\transformer(\latenttokenscoarse) \rightarrow [\querytokenscoarse; \keytokenscoarse]$ is decomposed into query tokens $\querytokenscoarse \inR{\numverticescoarse \times \modeldim}$ and $\keytokenscoarse \inR{\numgraphpoints \times \modeldim}$ (after separate linear projections), representing the refined template and feature tokens, respectively. 
The coarse vertices $\verticescoarse$ are then regressed as an attention-weighted sum of the sampling graph coordinates $\samplinggraph$: $\verticescoarse = \text{Softmax} (\querytokenscoarse \keytokenscoarse^{T}/\sqrt{\modeldim} ) \samplinggraph$.

\subsection{Shell-based prediction}

\paragraph{Feature sampling}
The coarse mesh $\meshcoarse$ is used to construct a sampling graph that layers the estimated surface.
Specifically, we compute surface layers displaced along the vertex normal directions.
Given the coarse predicted mesh $\meshcoarse$, we compute the vertex normals $\meshcoarsenormals \inR{\numverticescoarse \times 3}$.
We build a shell-based sampling point cloud by stacking the displaced vertices $\samplingshells = [\verticescoarse;\verticescoarse+\samplingshelldistance\meshcoarsenormals; \verticescoarse-\samplingshelldistance\meshcoarsenormals] \inR{\numshellpoints \times 3}$, where $\samplingshelldistance$ is the layer displacement distance.
Again, each sampling point $\samplingpoint \inR{3}$ of $\samplingshells$ is projected into each feature map $\{\featureimage_k\}_{k=1}^{n_i}$ to get per-view feature vectors $\featurevector_k \inR{\featuredim}$ via bilinear sampling.

\paragraph{Surface-aware feature fusion:}
To fuse these per-view features, we adopt the visibility-aware aggregation strategy of TEMPEH~\cite{bolkart2023tempeh}.
Unlike the graph stage which treats all views equally, this surface-aware fusion weights each view based on the local surface geometry of $\meshcoarse$.
For a sampling point $\samplingpoint$ of $\samplingshells$ associated with a vertex $\vertex$ of $\meshcoarse$, we compute a view-dependent weight $\featureweights_k = \text{Softplus}(\vertexvisibility_k \cdot \cos\theta_k$).
Here, $\vertexvisibility_k \in \{0, 1\}$ denotes the visibility of $\vertex$ from the $k$-th camera, determined via a depth-buffer check on the intermediate mesh $\meshcoarse$.
The term $\cos\theta_k=\vertexnormal^{T}\viewdir_k$ is the dot product between the vertex normal $\vertexnormal$ of $\vertex$ and the viewing direction $\viewdir_k = (\cameracenter_k-\vertex)/\norm{(\cameracenter_k-\vertex)}$, where $\cameracenter_k$ is the $k$-th camera center.
The Softplus function ensures positive weights and maintains non-zero gradients across all views.
Using these weights, we compute the weighted mean $\featuremean$ and weighted variance $\featurevariance$ across all $\numviews$ views, which are concatenated to form the fused feature vector $\featurevector = [\featuremean; \featurevariance]$.
Performing this feature fusion for all points in $\samplingshells$ yields the shell feature point cloud $\shellsfeaturepointcloud \inR{\numshellpoints \times \fusedfeaturedim}$.

\paragraph{Transformer-based mesh prediction}
In the second stage, we predict the full-resolution vertices $\verticesfinal \inR{\numvertices \times 3}$ by attending over the shell-based features.
The output resolution is established by the template vertices $\verticestemplate \inR{\numvertices \times 3}$, which are processed by an MLP to generate template tokens $\templatetokens \inR{\numvertices \times \modeldim}$.  

To obtain consistent semantic embeddings of the shell point cloud, we define fixed template shells $\samplingshellstemplate \inR{\numshellpoints \times 3}$ by displacing the downsampled template vertices $\verticestemplatedown$ along their vertex normals.
Unlike the dynamic sampling shells $\samplingshells$, which depend on the coarse prediction, this template shell is static and serves to encode the relative spatial relationships of the sampling points. 
The coordinates of $\samplingshellstemplate$ are similarly processed through an MLP to form geometric positional embeddings.
These embeddings are element-wise added to the fused multi-view features $\shellsfeaturepointcloud$ and projected to $\modeldim$ to produce shell feature tokens $\featuretokensshells \inR{\numshellpoints \times \modeldim}$.

Identical to the graph stage, the joint set of tokens $\latenttokensfinal = [\templatetokens; \featuretokensshells] \inR{(\numvertices+\numshellpoints) \times \modeldim}$ is processed by the transformer model $\transformer$, which is shared across the coarse and final stage.
The transformer output is partitioned and passed through separate linear layers, shared with the graph stage, to obtain query tokens $\querytokensfinal \inR{\numvertices \times \modeldim}$ and key tokens $\keytokensfinal \inR{\numshellpoints \times \modeldim}$.
The final vertices $\verticesfinal$ are then regressed as the attention-weighted sum of the dynamic sampling shell coordinates $\samplingshells$: $\verticesfinal = \text{Softmax} (\querytokensfinal \keytokensfinal^{T}/\sqrt{\modeldim}) \samplingshells$.

While ToFu~\cite{li2021tofu} and TEMPEH~\cite{bolkart2023tempeh} utilize $512$ ($8^3$) volumetric samples per vertex (totaling $9 \times 10^6$) for independent local refinement, \modelname regresses vertex positions as a weighted combination of only $9,000$ layered shell points.
By reducing the total number of 3D sampling locations across both stages to $11,592$, we drastically minimize the sampling operations. 
This significantly lowers GPU memory overhead and accelerates both training and inference.

\subsection{Loss functions}

We train \modelname end-to-end using a combination of vertex-to-vertex and point-to-plane distances between the reconstructed vertices and the ground-truth vertices. 
To supervise the first stage, the predicted coarse vertices $\verticescoarse$ are upsampled to the full mesh resolution using the barycentric upsampling matrix $\verticescoarseup = \upsamplingmatrix \verticescoarse \inR{\numvertices \times 3}$.
We define the displacement matrices for the upsampled coarse prediction $\verticescoarsedistance = \verticescoarseup - \verticesgt$ and the final predictions $\verticesfinaldistance = \verticesfinal - \verticesgt$.

\paragraph{Vertex-to-vertex (V2V) loss}
The V2V loss minimizes the Euclidean distances between predicted and ground truth vertex positions.
To allow for spatially varying importance across the mesh surface (e.g., to prioritize facial features), we introduce a diagonal weight matrix $\vertexweights = \text{diag}(\vertexweight_1, \dots, \vertexweight_{\numvertices})$, where $\vertexweight_i$ denotes the individual weight of the $i$-th vertex.
The loss is defined as:
\begin{equation}
    \lossvertextovertex = \lossweightcoarse \norm{\vertexweights \verticescoarsedistance}_F^2 + \lossweightfinal \norm{\vertexweights \verticesfinaldistance}_F^2,
\end{equation}
where $\lossweightcoarse$ and $\lossweightfinal$ balance the coarse and final prediction stages.

\paragraph{Vertex-to-plane loss:}
To improve surface alignment, we incorporate a vertex-to-plane loss, formulated as:
\begin{equation}
    \lossvertextoplane = \lossweightcoarse \norm{\vertexweights (\verticescoarsedistance \odot \meshgtnormals) \mathbf{1}_3}_2^2 + \lossweightfinal \norm{\vertexweights (\verticesfinaldistance \odot \meshgtnormals) \mathbf{1}_3}_2^2,
\end{equation}
where $\odot$ denotes the Hadamard product, and $\mathbf{1}_3 \inR{3}$ is a column vector of ones used to perform row-wise summation, effectively computing the dot product between displacement vectors and ground-truth vertex normals $\meshgtnormals \inR{\numvertices \times 3}$. 
This loss penalizes only the displacement component orthogonal to the target surface, which allows vertices to distribute along the geometry by avoiding penalties for "sliding" along the local tangent planes.

\paragraph{Total loss}
The model is trained by minimizing:
\begin{equation}
    \losstotal = \lossweightvertextovertex \lossvertextovertex + \lossweightvertextoplane \lossvertextoplane,
\end{equation}
with weights $\lossweightvertextovertex$ and $\lossweightvertextoplane$ of the individual losses.

Optimization-based frameworks~\cite{egger2020survey} and TEMPEH~\cite{bolkart2023tempeh} typically minimize point-to-surface (P2S) distances, which permit tangential sliding and mesh distortions that necessitate explicit regularization.
In contrast, \modelname leverages a vertex-to-vertex (V2V) loss based on dense semantic correspondence.
This provides strong implicit regularity, maintaining surface integrity without requiring additional regularization.

\section{Implementation details}
\label{sec:implementation}

\begin{figure}[htp]
    \centering
    \includegraphics[width=1.0\columnwidth]{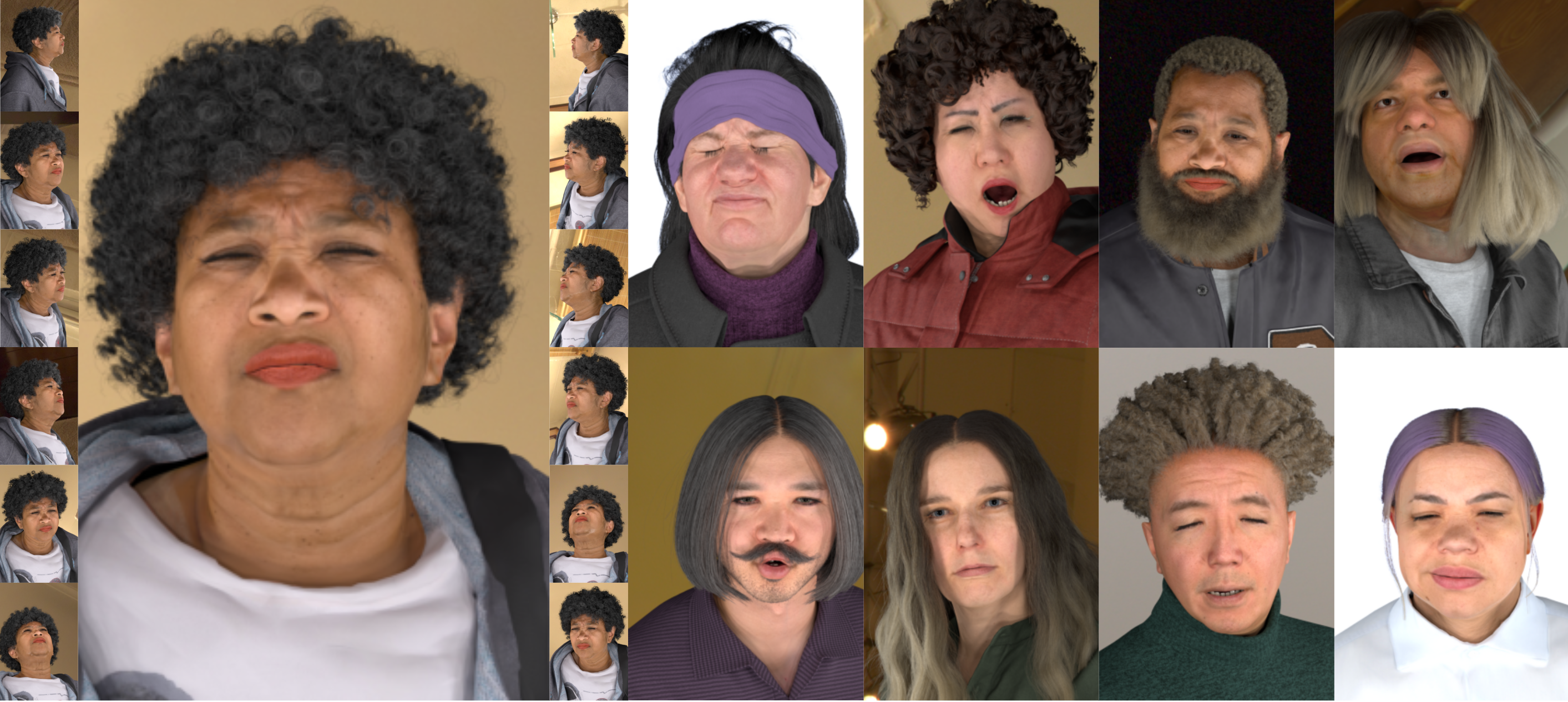}  
    \caption{
        \textbf{Synthetic dataset.}
        (Left) A single subject rendered from 13 camera views simulating a multi-view capture environment.
        (Right) Random samples demonstrating the diversity in identities and expressions, augmented with randomized backgrounds and assets including clothing and hair.
    }
	\label{fig:dataset}
\end{figure}

\paragraph{Synthetic dataset}
We adopt the procedural approach of \citet{wood2021fakeit} to construct a synthetic dataset (see Fig.~\ref{fig:dataset}) with paired calibrated multi-view images and meshes in a unified mesh topology.
First, we select a mesh from an internal dataset with registered 3D head meshes (with $17,821$ vertices each) of $\geq 2500$ identities.
Each mesh is assigned a skin texture and a randomized blend of facial expressions.
To increase the diversity and realism of the training data, these textured meshes are augmented with various assets including clothing, facial and scalp hair, and accessories.
We render the scenes using Blender's Cycles engine from $13$ predefined camera views, with each render composited over a randomly selected background.
Each image is rendered with resolution $1536 \times 1024$.
The camera configuration covers the frontal hemisphere of the face, simulating our physical multi-view capture environment.
In total, the dataset consists of $300,000$ data pairs across $2,064$ unique identities (varying in face shape and appearance).
The identities represent these demographics: $46\%/54\%$ female/male; age groups 20--35 ($48\%$), 36--55 ($39\%$), and 56+ ($12\%$); and ethnicities comprising White ($38\%$), East Asian ($24\%$), South Asian ($12\%$), Hispanic/Latino ($9\%$), Black, ($9\%$), Southeast Asian ($5\%), $Middle Eastern ($4\%$), and others ($1\%$).

\paragraph{Training data}
We partition the synthetic dataset into training, validation, and test sets using an $80\%/10\%/10\%$ ratio, ensuring disjoint identities across splits.
In total, the training (validation) data consists of $251,816$ ($27,295$) samples across $1,674$ ($181$) identities.
For training, input images are downsampled by a factor of $4$ to an effective resolution of $384 \times 256$.

\paragraph{Parameter settings:}
The feature extractor $\featurenet$ utilizes a DINOv2-B backbone \cite{oquab2023dinov2}  with four registers and LoRA \cite{hu2022lora} residuals ($\lorarank = 5$).
We concatenate four evenly spaced backbone layers and project them to $\featuredim = 98$ via $1\times1$ convolutions and pixel shuffling to output feature maps with $\featureimagewidth = 27$ and $\featureimageheight = 18$ for the $384 \times 256$ input.
The shared transformer $\transformer$ uses an XCiT architecture following the ViT-S configuration ($12$ layers, $6$ heads, $\modeldim = 384$) with each layer's FFN using a $1,536$-dimension inner layer and GELU activation.
For prediction, the graph stage ($\numverticescoarse=3,000$) employs a sampling graph $\samplinggraph$ of $16$ concentric shells of twice-subdivided icosahedra with $25\text{mm}$ radial spacing, totaling $\numgraphpoints=2,592$ points. 
The final stage uses surface-aware shells $\samplingshells$ displaced at $\pm 4\text{mm}$ (i.e., $\samplingshelldistance = 4$) around $\verticescoarse$ ($\numshellpoints=9,000$ points).
Template vertices are tokenized via a shared MLP consisting of three linear layers ($2\modeldim$ projection) with GELU activations and a final linear projection to $\modeldim$.
The sampling points for positional embeddings ($\samplinggraph, \samplingshellstemplate$) are processed by two linear layers with an intermediate GELU activation.

The model is implemented in PyTorch~\cite{paszke2019pytorch} and optimized using AdamW~\cite{loshchilov2019adamw} with a batch size of $3$ for $900,000$ steps.
Training takes approximately 2 weeks on a single NVIDIA H100 80GB HBM3 GPU.
We use a $10$k-step linear warmup to a $1\text{e-}4$ learning rate, held constant for $100,000$ steps, followed by exponential decay.
Training is phased: the first $500$k iterations focus on the coarse and LoRA layers ($\lossweightcoarse=1.0, \lossweightfinal=0.0$), after which the full model is trained end-to-end ($\lossweightfinal=1.0$).
Vertex-to-vertex and vertex-to-plane losses are weighted equally ($\lossweightvertextovertex = \lossweightvertextoplane = 1.0$), with vertex weights $\vertexweights$ prioritizing facial features ($5.0$ for lips/eyelids, $3.0$ for skin, eyebrows, ears, and nose), and $1.0$ for the rest (e.g., mouth interior, teeth, neck, scalp).

Data augmentation includes independent per-view color augmentations (brightness $\pm0.2$, contrast/saturation $0.9$--$1.1$, hue $\pm0.02$) and  global geometric transformations ($\pm 45^\circ$ rotation, $0.9$--$1.4\times$ scaling), with camera intrinsics updated accordingly. 
To ensure camera robustness, we randomly sample $8$--$13$ views per batch and apply random rotations to the coarse sampling graph.

\section{Evaluation}
\label{sec:evaluation}

\begin{figure*}[htp]
    \centering
    \includegraphics[width=1.0\textwidth]{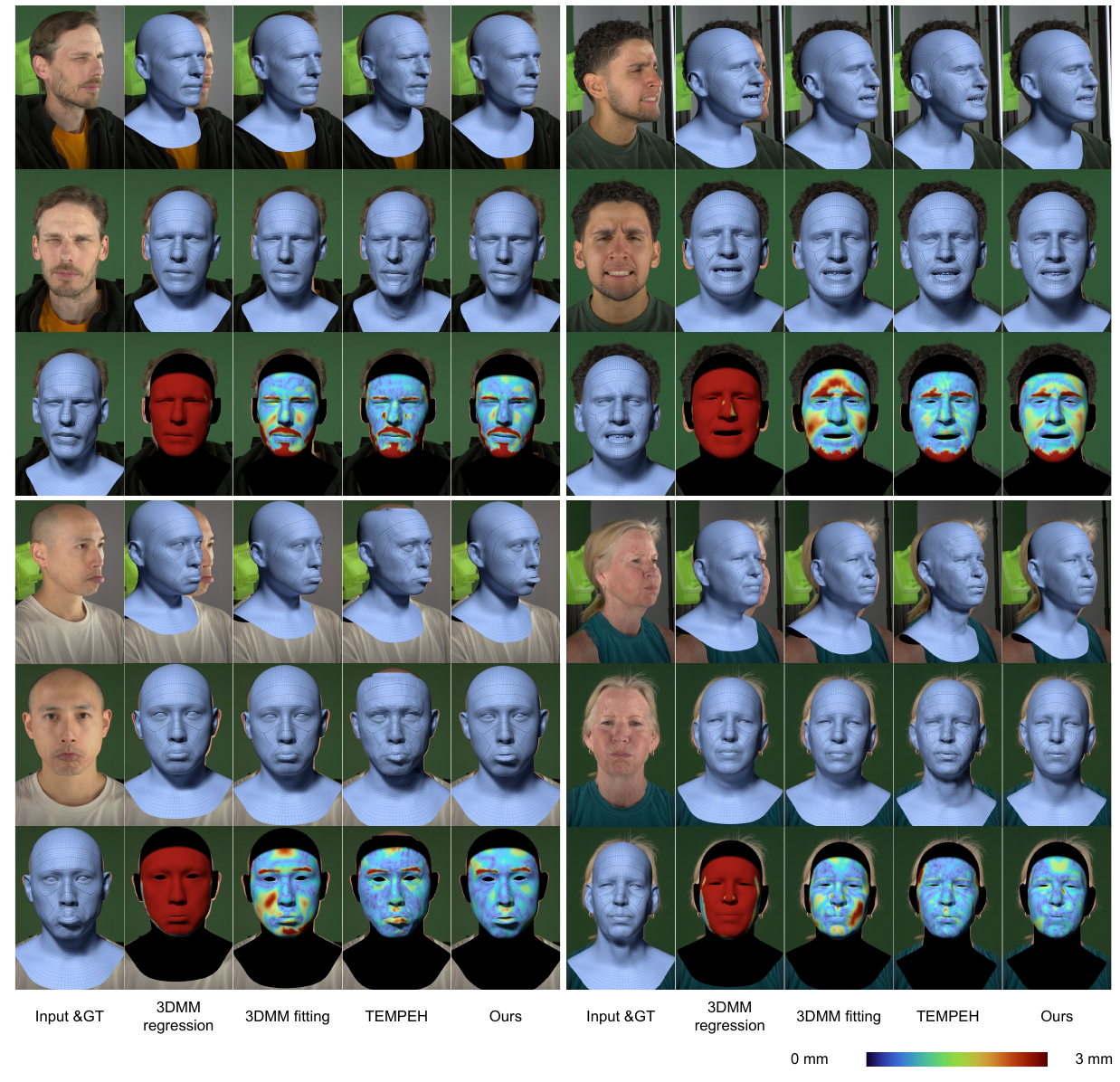}  
    \caption{
        \textbf{Baseline comparisons.}
        Comparison to the 3DMM regression, 3DMM fitting \cite{wood2021fakeit}, and TEMPEH \cite{bolkart2023tempeh}. 
        For each sample, we show one side view, a frontal view, and a rendering of the reference registration overlaid with the frontal image.
        The error visualizes the color coded (range $0-3$ mm) point-to-surface distance of each point in the reconstructed mesh and the closest point in the surface of the reference scan. 
    }
	\label{fig:qualcompare}
\end{figure*}
\begin{figure*}[htp]
    \centering
    \includegraphics[width=1.0\textwidth]{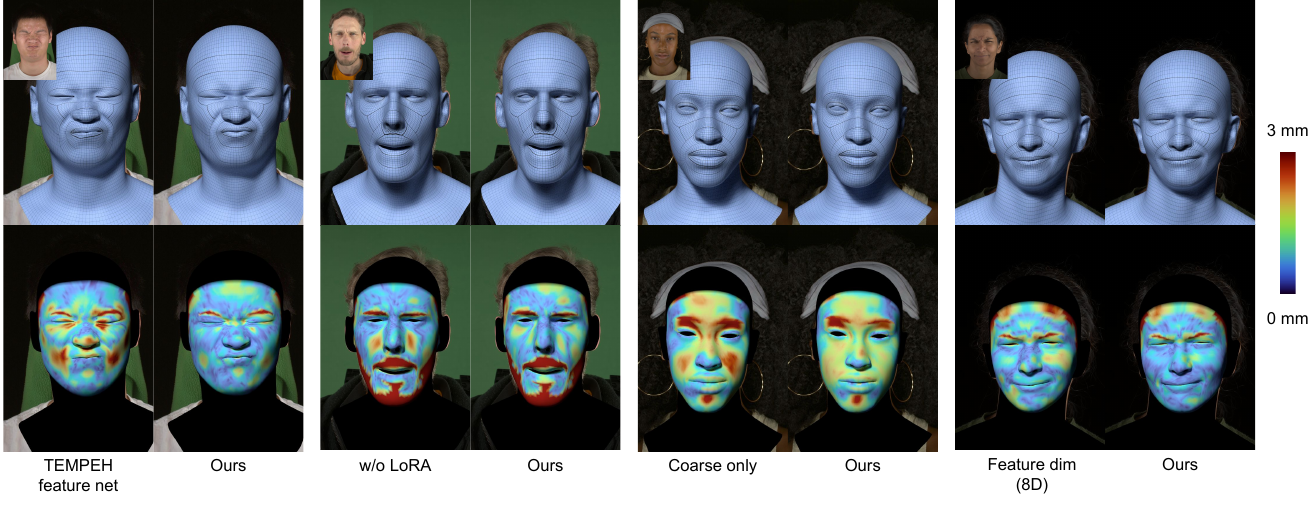}  
    \caption{
        \textbf{Ablations.}
        We show qualitative comparisons of \modelname (Ours) to different ablated model variants. 
    }
	\label{fig:ablation}
\end{figure*}

\paragraph{Test data.}
\modelname is evaluated on held-out synthetic data and real-world multi-view capture. 
\textbf{(1)} \textit{Synthetic test.} This set comprises $30,889$ procedural samples across $209$ identities disjoint from the training and validation sets.
\textbf{(2)} \textit{Real test.} We utilize $9,617$ multi-view frames from $303$ subjects recorded in approximately $50$ static expressions using a calibrated $13$-camera system.
The system provides full camera parameters, including extrinsics, intrinsics, and lens distortions.
To establish reference geometry, we reconstruct unstructured scans via deep MVS \cite{qiu2024chosen} and register them by fitting a 3DMM, guided by predicted dense per-view landmarks, followed by non-rigid surface deformation.
Both raw scans and registrations are used for evaluation.

\paragraph{Baselines.}
We compare \modelname against TEMPEH~\cite{bolkart2023tempeh}, a 3DMM regressor, and a traditional multi-view fitting.

\textbf{(1)} \textit{TEMPEH}.
TEMPEH performs feed-forward multi-view registered mesh inference.
We train it on our synthetic data for $1.4$M steps ($\sim$30 days on an NVIDIA H100) using a batch size of one due to its high training memory demand ($\sim$65GB).
To ensure a fair comparison, we adapted the original implementation to use the same supervision, loss functions, and per-vertex weighting as \modelname.
Following our schedule, we optimized its global stage for 500k iterations before training end-to-end.

\textbf{(2)} \textit{3DMM regressor.}
We implement a baseline that regresses $116$ identity and $197$ expression shape parameters of a 3DMM.
The model combines linear blend skinning for neck and eyeball articulation with linear identity and expression blend shapes, following existing models \cite{li2017flame, wood2021fakeit, bednarik2024stabilize}.
For each multi-view input, we detect dense landmarks \cite{wood2022denselandmarks}, crop the facial region, and process each image through a monocular 3DMM parameter regressor.
The final multi-view result is obtained by averaging the predicted parameters across all views.

\textbf{(3)} \textit{Multi-view fitting.}
We jointly fit the 3DMM to dense per-view landmarks, following \cite{wood2022denselandmarks}, taking around $35$ seconds.

\subsection{Qualitative evaluation}

Figure~\ref{fig:qualcompare} qualitatively compares \modelname with 3DMM regression, 3DMM fitting \cite{wood2021fakeit}, and TEMPEH \cite{bolkart2023tempeh}.
The 3DMM regression baseline generally lacks the metric accuracy required for multi-view reconstruction.
While the 3DMM fitting achieves better alignment via per-view landmark guidance, it results in poor fits in the cheek and forehead regions (Fig.~\ref{fig:qualcompare}, top right) and fails to capture subtle dynamics such as lip rolling (bottom left) or cheek blowing (bottom right) as effectively as our method.
TEMPEH recovers more geometric detail and fits the scan surface more closely, as is reflected in the error visualizations, however, it lacks global surface coherence and frequently produces noisy mesh estimates (see Figure~\ref{fig:noisytempeh}).
Instead, \modelname consistently produces smooth and visually coherent meshes.

\subsection{Quantitative evaluation}
\begin{table}[t]
    \centering
    \footnotesize
    \caption{
        \textbf{Ablations and comparisons on synthetic data.}
        Ablations of individual model components and TEMPEH \cite{bolkart2023tempeh} baseline comparisons.
        We report the vertex-to-vertex (V2V) ground truth distance on the face region (left) and the entire mesh (right).
    }
    \label{tab:evaluation_synthetic_data}
    \setlength{\tabcolsep}{3pt}
    \begin{tabular}{l ccc ccc}
    \toprule
    \multirow{2.5}{*}{\textbf{Method}} & \multicolumn{3}{c}{\textbf{V2V (mm) - face only}} & \multicolumn{3}{c}{\textbf{V2V (mm) - full mesh}}\\
    \cmidrule(lr){2-4} \cmidrule(lr){5-7}
     & Mean & Median & Std & Mean & Median & Std \\
    \midrule
    
    \textbf{Feature extraction} & & & & & &\\
    \hspace{2mm} (1) TEMPEH feature net & 2.00 & 1.82 & 1.06 & 2.49 & 2.24 & 1.38\\
    \hspace{2mm} (2) W/o downsampling & 1.97 & 1.65 & 1.23 & 2.41 & 2.12 & 1.41\\
    \hspace{2mm} (3) Feature dim. (8D) & 1.56 & 1.39 & 0.85 & 2.03 & 1.79 & 1.18\\
    \hspace{2mm} (4) W/o LoRA & 1.62 & 1.46 & 0.86 & 2.11 & 1.87 & 1.21\\
    \midrule
    \textbf{Mesh prediction} & & & \\
    \hspace{2mm} (5) Samp. density (w/o subdiv) & 1.85 & 1.65 & 1.01 &2.57 & 2.27 & 1.53\\
    \hspace{2mm} (6) Samp. density (1 subdiv) & 1.56 & 1.39 & 0.86 & 2.20 & 1.94 & 1.31\\
    \hspace{2mm} (7) Coarse res. (500) & 1.43 & 1.26 & 0.82  & 1.93 & 1.66 & 1.20 \\
    \hspace{2mm} (8) Coarse only & 1.50 & 1.32 & 0.83 & 2.00 & 1.76 & 1.19 \\
    
    \midrule
    \textbf{TEMPEH} & & & & & & \\
    \hspace{2mm} Global  & 2.09 & 1.87 & 1.11 & 2.67 & 2.36 & 1.55\\
    \hspace{2mm} Refinement  & 1.94 & 1.71 & 1.09 & 2.59 & 2.25 & 1.56 \\
    \midrule
    \textbf{\modelname (ours)} & & & & & & \\
    \hspace{2mm} Coarse  & 1.47 & 1.30 & 0.83 & 1.92 & 1.66 & 1.25 \\
    \hspace{2mm} Final  & 1.39 & 1.22 & 0.79 & 1.83 & 1.59 & 1.12 \\
    \bottomrule
    \end{tabular}
\end{table}
\begin{table}[t]
    \centering
    \footnotesize
    \caption{
        \textbf{Comparisons on capture data.}
        Baseline comparisons against 3DMM regression, 3DMM fitting \cite{wood2022denselandmarks}, and TEMPEH \cite{bolkart2023tempeh}.
        We report the vertex-to-vertex (V2V) distance to registrations and the point-to-surface (P2S) distance between MVS scans and the predictions.
        V2V 3DMM fitting results are in brackets, as the registrations use the fitting results for initialization, creating an inherent bias.
    }
    \label{tab:comparison_real_data}
    \setlength{\tabcolsep}{3pt}
    \begin{tabular}{l ccc ccc}
    \toprule
    \multirow{2.5}{*}{\textbf{Method}} & \multicolumn{3}{c}{\textbf{V2V (mm)}} & \multicolumn{3}{c}{\textbf{P2S (mm)}} \\
    \cmidrule(lr){2-4} \cmidrule(lr){5-7}
     & Mean & Median & Std & Mean & Median & Std \\
    \midrule
    \textbf{3DMM regression} & 30.33 & 30.20 & 1.52 & 17.23 & 18.31 & 9.37\\
    \midrule
    \textbf{3DMM fitting}  & (1.53) & (1.33) & (0.94)  & 2.05 & 1.03 & 3.31\\
    \midrule
    \textbf{TEMPEH}  & & & & & &\\
    \hspace{2mm} Coarse  & 2.28 & 2.06 & 1.19 & 1.66 & 1.09 & 2.27 \\
    \hspace{2mm} Final   & 2.13 & 1.90 & 1.16 & 1.19 & 0.62 & 2.10 \\
    \midrule
    \textbf{\modelname (ours)}  & & & & & & \\
    \hspace{2mm} Coarse   & 1.74 & 1.53 & 1.00 & 1.14 & 0.78 & 1.17 \\
    \hspace{2mm} Final   & 1.71 & 1.50 & 0.97 & 1.13 & 0.76 & 1.17 \\
    \bottomrule
    \end{tabular}
\end{table}

\paragraph{Data and metrics.}
We evaluate reconstruction accuracy on the synthetic test set (\syntheticdata) and the real capture test set (\realdata) using four metrics.
On \syntheticdata, providing topologically consistent ground truth, we compute Euclidean vertex-to-vertex (V2V) distances.
Since \realdata lacks ground truth, we report V2V distances relative to reference registrations.
For geometric fidelity against MVS scans, we compute point-to-surface (P2S) distances between skin-labeled scan vertices and the predicted mesh, where skin labels are obtained by projecting 2D face parsing results onto the 3D scans. 
While V2V assesses semantic correspondence accuracy, P2S evaluates the geometric fidelity relative to the unstructured MVS data.
We also quantify mesh quality through triangle distortion and orientation.
Distortion is measured as the mean absolute difference in triangle area between prediction and the ground truth. 
To detect mesh artifacts, we evaluate orientation by identifying inverted normals.
We report this as a percentage of triangles per mesh to indicate the frequency of triangle flips.

\paragraph{Baseline comparison.}
As shown in Tab.~\ref{tab:evaluation_synthetic_data}, \modelname outperforms TEMPEH~\cite{bolkart2023tempeh} on synthetic data, achieving a $29\%$ ($28\%$) lower median (mean) V2V error.
This performance gain generalizes to capture data (Tab.~\ref{tab:comparison_real_data}), with a $21\%$ ($20\%$) median (mean) V2V error reduction compared to TEMPEH.
A notable distinction arises in the P2S metrics on capture data.
While \modelname achieves a $5\%$ lower mean P2S error, TEMPEH exhibits an $18\%$ lower median P2S distance.
This indicates that TEMPEH's per-vertex refinement effectively pulls individual points toward the scan boundary, but at the cost of global surface coherence.
The higher V2V error for TEMPEH confirms that while its vertices may lie closer to the scan surface, they do not maintain accurate semantic correspondence.
In contrast, \modelname's holistic prediction ensures superior correspondence across the entire head.
For the baseline comparisons on capture data (Tab.~\ref{tab:comparison_real_data}), the 3DMM regressor performs poorly, primarily because the monocular parameter estimation lacks metric accuracy and simple averaging across views fails to resolve depth ambiguities, leading to high V2V errors.
Furthermore, 3DMM fitting results in a $36\%$ ($81\%$) higher median (mean) P2S error compared to \modelname, indicating an overall worse face geometry reconstruction.

Quantifying mesh quality, \modelname significantly outperforms TEMPEH, achieving a $31\%$ lower triangle deformation score ($0.38$ vs. $0.55$) and nearly half the triangle flip rate ($0.08\%$ vs. $0.15\%$).
\modelname performs similarly to the 3DMM regressor across both metrics (deformation: $0.44$, orientation: $0.08\%$).
The optimization-based 3DMM fitting yields lower scores (deformation: $0.29$, orientation: $0.07\%$), but requires orders of magnitude more computation time.

Note that the reference-based scores for the 3DMM fitting are inherently biased.
The reference registrations used for evaluation are initialized via the 3DMM fitting process and utilize the same dense landmarks during optimization.
Consequently, the reference geometry is predisposed toward the fitting baseline.

\subsection{Ablation experiments}
\begin{table}[t]
    \centering
    \footnotesize
    \caption{
        \textbf{Ablations on capture data.} 
        We evaluate the contribution of individual components in the feature extraction and mesh prediction stages.
        We report the vertex-to-vertex (V2V) distance to registrations and the point-to-surface (P2S) distance between MVS scans and the predictions.}
    \label{tab:ablation_real_data}
    \setlength{\tabcolsep}{3pt}
    \begin{tabular}{l ccc ccc}
    \toprule
    \multirow{2.5}{*}{\textbf{Method}} & \multicolumn{3}{c}{\textbf{V2V (mm)}} & \multicolumn{3}{c}{\textbf{P2S (mm)}} \\
    \cmidrule(lr){2-4} \cmidrule(lr){5-7}
     & Mean & Median & Std & Mean & Median & Std \\
    \midrule
    \textbf{\modelname (Ours)} & 1.71 & 1.50 & 0.97 & 1.13 & 0.76 & 1.17 \\
    \midrule
    \textbf{Feature extraction} & & & & & & \\
    \hspace{2mm} (1) TEMPEH feature net & 2.31 & 2.08 & 1.23 & 1.70 & 1.12 & 2.35 \\
    \hspace{2mm} (2) W/o downsampling & 2.22 & 1.92 & 1.28 & 1.34 & 0.96 & 1.29 \\
    \hspace{2mm} (3) Feature dim. (8D) & 1.80 & 1.59 & 1.01 & 1.25 & 0.86 & 1.28 \\
    \hspace{2mm} (4) W/o LoRA & 1.82 & 1.62 & 0.98 & 1.27 & 0.86 & 1.27 \\
    \midrule
    \textbf{Mesh prediction} & & & & & & \\
    \hspace{2mm} (5) Samp. density (w/o subdiv) & 2.07 & 1.85 & 1.13 & 1.54 & 1.11 & 1.49 \\
    \hspace{2mm} (6) Samp. density (1 subdiv) & 1.79 & 1.61 & 0.95 & 1.25 & 0.84 & 1.27 \\
    \hspace{2mm} (7) Coarse res. (500) & 1.71 & 1.50 & 0.97 & 1.18 & 0.80 & 1.20 \\
    \hspace{2mm} (8) Coarse only & 1.77 & 1.58 & 0.95 & 1.17 & 0.81 & 1.17 \\

    \bottomrule
    \end{tabular}
\end{table}

We evaluate our architectural design choices through ablation experiments on the synthetic test set (\syntheticdata) and the capture test set (\realdata), as summarized in Table~\ref{tab:evaluation_synthetic_data} and Table~\ref{tab:ablation_real_data}.
Unless otherwise noted, our discussion focuses on the median errors and face-only V2V metrics.

\paragraph{Feature extraction.}
We first evaluate the impact of the image feature extractor through several modifications: 
\textbf{(1)} \textit{TEMPEH feature net}: Replacing our DINOv2-based feature extractor with the ResNet34-UNet architecture used in TEMPEH \cite{bolkart2023tempeh} increases V2V by $49\%$ on \syntheticdata and $39\%$ on \realdata, while P2S increases by $47\%$.
This validates the superior geometric cues provided by foundation model features. 
\textbf{(2)} \textit{W/o downsampling}: Upsampling DINOv2 features to the original input resolution (8D output) via linear layers and pixel shuffling increases V2V by $35\%$ on \syntheticdata and $28\%$ on \realdata, while P2S increases by $26\%$.
This suggests that the native transformer feature resolution is more robust for surface alignment.
\textbf{(3)} \textit{Feature dimension (8D)}: Reducing the feature dimensionality to 8D (following TEMPEH) degrades performance, increasing V2V by $14\%$ on \syntheticdata and $6\%$ on \realdata, while P2S increases by $13\%$.
This indicates that lower-dimensional features lack the necessary detail for dense surface regression.
\textbf{(4)} \textit{W/o LoRA}: Freezing the backbone without task-specific adaptation leads to a $20\%$ V2V increase on \syntheticdata and $8\%$ on \realdata, while P2S increases by $13\%$, demonstrating the importance of the LoRA layers in adapting the general-purpose backbone to facial geometry reconstruction.

\paragraph{Mesh prediction.}
We further ablate the transformer-based prediction stages:
\textbf{(5 $\&$ 6)} \textit{Samp. density (w/o subdiv / 1 subdiv)}: Varying the resolution of the coarse sampling graph $\samplinggraph$ (using 192 or 672 points) demonstrates that the initial sampling density is critical.
With lower density (w/o subdivision), V2V increases by $35\%$ on \syntheticdata and $23\%$ on \realdata, while P2S increases by $46\%$.
Even with a single subdivision, V2V is $14\%$ higher on \syntheticdata and $7\%$ higher on \realdata, with a $5\%$ P2S increase.
\textbf{(7)} \textit{Coarse res. (500)}: Reducing the intermediate mesh to $\numverticescoarse = 500$ vertices marginally increases V2V on \syntheticdata by $3\%$, while yielding identical V2V on \realdata and a $5\%$ P2S increase.
This indicates that while the intermediate mesh guides the sampling shells, the final stage is robust to the specific resolution of the intermediate mesh.
\textbf{(8)} \textit{Graph stage only}: Omitting the second stage and regressing all $17,821$ vertices directly from the global graph results in an $8\%$ V2V increase on \syntheticdata, a $5\%$ increase on \realdata, and a $7\%$ P2S increase.
As shown in Fig.~\ref{fig:ablation}, while the coarse-only stage captures the general head shape, it lacks the fine surface detail recovered by our hierarchical shell-based refinement.

Finally, we find that training with an additional Laplacian~\cite{taubin1995laplacian} loss (with a weight of 1.0) yields no accuracy gains.
V2V errors (face only) remain identical on the synthetic test set (mean/median/std: 1.39/1.22/0.79 mm).
Results on capture test data are similarly unchanged, with V2V errors of 1.71/1.49/0.97 mm and P2S errors of 1.14/0.76/1.18 mm (mean/median/std).

\section{Discussion}
\label{sec:discussion}

\begin{figure*}[htp]
    \centering
    \begin{minipage}[b]{0.49\textwidth}
        \centering
        \includegraphics[width=\linewidth]{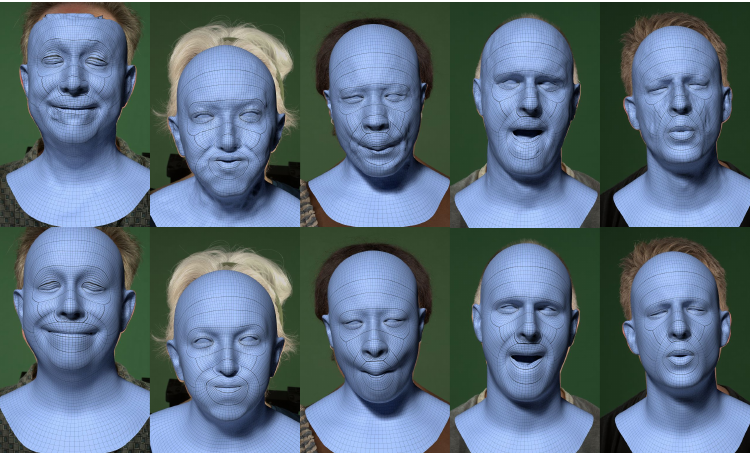}
        \captionof{figure}{
            \textbf{Mesh consistency.} 
            TEMPEH \cite{bolkart2023tempeh} (top) exhibits significant surface noise and artifacts, whereas \modelname (bottom) maintains global surface consistency and smoothness across various subjects.
        }
        \label{fig:noisytempeh}
    \end{minipage}
    \hfill
    \begin{minipage}[b]{0.49\textwidth}
        \centering
        \includegraphics[width=\linewidth]{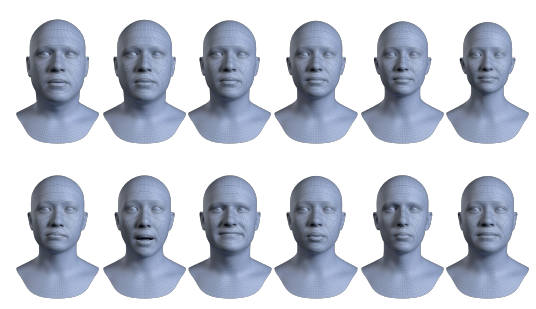}%
        \captionof{figure}{
            \textbf{Application to 3DMM building.} 
            Top row: \modelname simplifies the generation of registered meshes allowing us to easily build statistical 3DMMs of faces.
        Bottom row: We sample this 3DMM built from \modelname outputs to generate novel shapes and expressions.
        }
        \label{fig:pca}
    \end{minipage}
\end{figure*}

\begin{figure*}[htp]
    \centering
    \begin{minipage}[b]{0.49\textwidth}
        \centering
        \includegraphics[width=\linewidth]{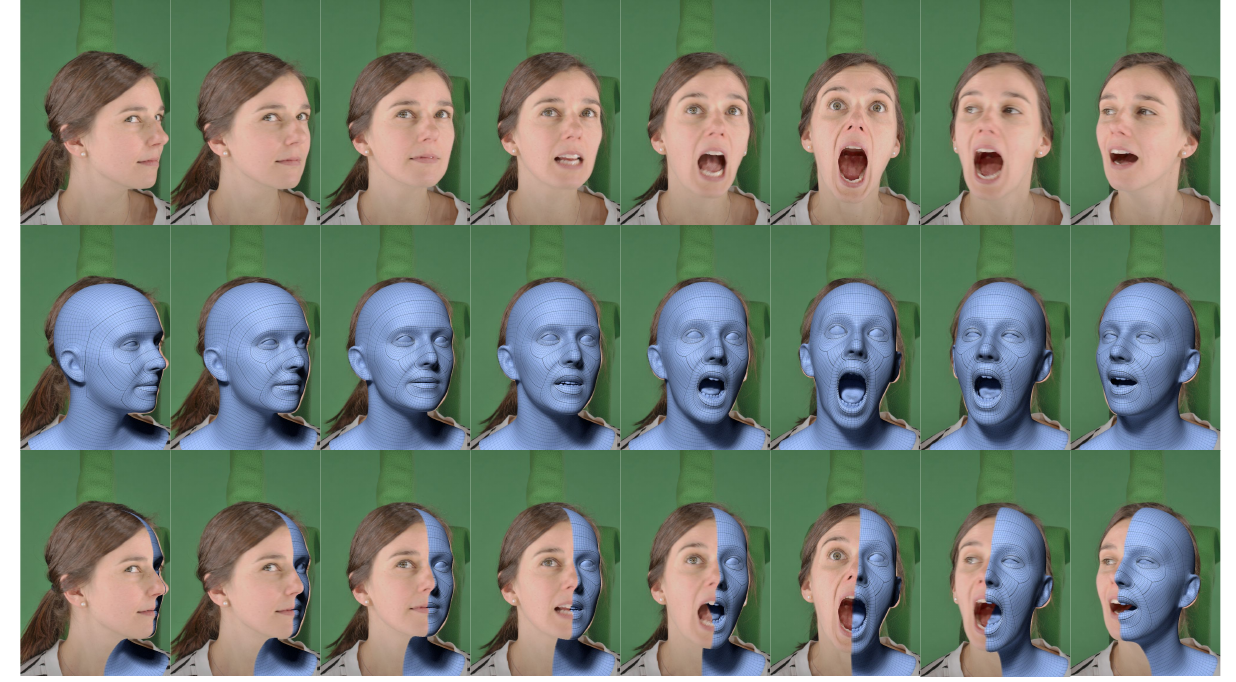}
        \captionof{figure}{
            \textbf{Performance registration.} 
            \modelname can be applied frame-by-frame to dynamic facial performances and produces temporally smooth and expressive performance registrations.
            See the video for the full performances.
        }
        \label{fig:perfreg}
    \end{minipage}
    \hfill
    \begin{minipage}[b]{0.50\textwidth}
        \centering
        \includegraphics[width=\linewidth]{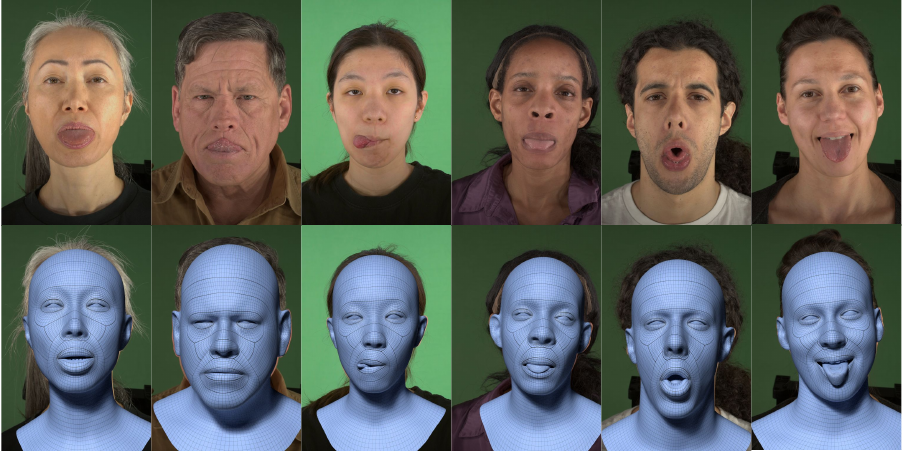}%
        \captionof{figure}{
            \textbf{Failure cases.} 
            Our framework occasionally struggles with extreme tongue articulations. This is primarily attributed to the limited diversity of tongue expressions within our synthetic training set.
        }
        \label{fig:failures}
    \end{minipage}
\end{figure*}

\begin{figure*}[htp]
    \centering
    \includegraphics[width=1.0\textwidth]{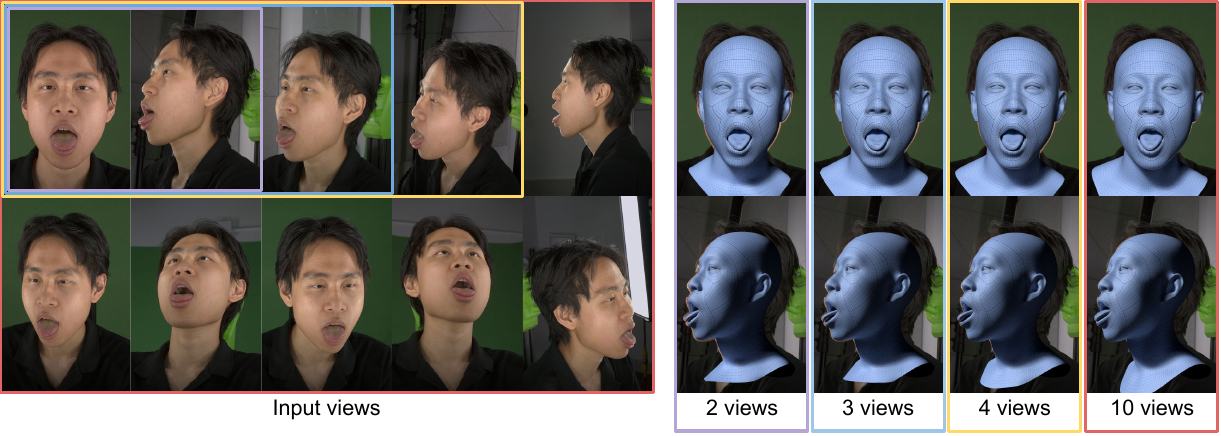}  
    \caption{
        \textbf{Varying views at inference.}
        \modelname is robust to the number of input views.
        Here we show predictions given 2, 3, 4, and 10 input views for the same subject.        
        Our predictions remain plausible even with just 2 input views featuring large disparities that challenge traditional MVS methods.
    }
	\label{fig:viewablation}
\end{figure*}

\paragraph{Applications}
\modelname bypasses traditional registration, enabling efficient 3DMM construction after rigid stabilization~\cite{bednarik2024stabilize} (Fig.~\ref{fig:pca}).
It further facilitates large-scale performance capture, yielding temporally consistent reconstructions even when applied frame-by-frame, without requiring temporal filtering or post-processing.
Figure~\ref{fig:perfreg} displays predictions for sample frames, while the supplementary video highlights the temporal stability of the results for multiple subjects and facial performances by visualizing the shared mesh topology across entire sequences.

\paragraph{Limitations}
\modelname fails to reconstruct certain tongue expressions (Fig.~\ref{fig:failures}) due to limited synthetic training diversity.
Improving these requires more varied tongue training configurations.

\paragraph{Detail reconstruction}
The predicted 18k-vertex meshes capture global structure and mid-frequency features but lack geometric details (e.g., fine wrinkles and skin pores) required for photorealistic rendering.
To achieve higher realism, a separate synthesis network as in ToFu~\cite{li2021tofu} could be trained to predict displacement maps and textures atop our output.

\paragraph{Occlusion}
Through global attention, \modelname handles occlusions (e.g.,  hair, mouth cavity) by correlating visible areas with a learned geometric prior to regress all 18k vertices holistically with a fixed mesh connectivity.
Interior mouth vertices are implicitly tucked into the cavity during closure to maintain semantic consistency without adding edges between the lips even if vertices coincide.

\paragraph{Modeling non-skin surfaces}
\modelname is optimized to predict the skin surface beneath hair or clothing.
However, neural avatars~\cite{zielonka2025synshot, qian2024gaussianavatars} often require mesh proxies aligned with the outer volume of hair or beards.
Reconstructing these holistic volumes requires extending our synthetic dataset to include consistent hair and clothing surface labels.

\paragraph{Number of input views}
Random camera dropout during training and mean-variance feature fusion make \modelname robust to varying input image counts.
While single-view reconstruction is ill-posed, results are reasonable with as few as two views (Fig.~\ref{fig:viewablation}).
\section{Conclusion}
\label{sec:conclusion}

We have presented \modelname, an efficient feed-forward framework for 3D head reconstruction in dense semantic correspondence from calibrated multi-view images.
The core of our approach lies in a hierarchical strategy that combines a sparse global sampling graph with dynamic, surface-aware sampling shells.
By decoupling feature extraction from the final mesh resolution, this design enables the model to scale to high-resolution topologies ($\geq$18k vertices) while requiring only $12\%$ of the GPU memory used by previous volumetric approaches.
Furthermore, by replacing independent per-vertex refinement with a holistic transformer-based prediction, \modelname maintains global surface consistency and demonstrates superior robustness to occlusions.
Notably, the model generalizes effectively from synthetic training to real-world captures, eliminating the need for the costly pre-registered multi-view datasets required by prior work.
Experimentally, \modelname achieves a median registration error $21\%- 29\%$ lower than the previous state-of-the-art on both real and synthetic data.
With its combination of geometric accuracy and sub-second inference speed, \modelname provides a scalable solution for real-time multi-view performance capture.

\begin{acks}

We thank M. Prinzler and V. Choutas for their helpful discussions and proofreading, D. Vicini for assistance with Mitsuba rendering, and E. Wood for support with synthetic data generation.

\end{acks}

\balance
\bibliographystyle{ACM-Reference-Format}
\bibliography{bibliography}

\end{document}